# Localization-Guided Track: A Deep Association Multi-Object Tracking Framework Based on Localization Confidence of Detections

Ting Meng[1]    Chunyun Fu[1]    Mingguang Huang[1]    Xiyang Wang[1]    Jiawei He[1]
Tao Huang[2]    Wankai Shi[1]

[1]Chongqing University    [2]James Cook University

mengting@stu.cqu.edu.cn, fuchunyun@cqu.edu.cn, huangmingguang@stu.cqu.edu.cn,
wangxiyang@cqu.edu.cn, hejiawei@cqu.edu.cn, tao.huang1@jcu.edu.au, wankai_shi@cqu.edu.cn

## Abstract

*In currently available literature, no tracking-by-detection (TBD) paradigm-based tracking method has considered the localization confidence of detection boxes. In most TBD-based methods, it is considered that objects of low detection confidence are highly occluded and thus it is a normal practice to directly disregard such objects or to reduce their priority in matching. In addition, appearance similarity is not a factor to consider for matching these objects. However, in terms of the detection confidence fusing classification and localization, objects of low detection confidence may have inaccurate localization but clear appearance; similarly, objects of high detection confidence may have inaccurate localization or unclear appearance; yet these objects are not further classified. In view of these issues, we propose* Localization-Guided Track *(LG-Track). Firstly, localization confidence is applied in MOT for the first time, with appearance clarity and localization accuracy of detection boxes taken into account, and an effective deep association mechanism is designed; secondly, based on the classification confidence and localization confidence, a more appropriate cost matrix can be selected and used; finally, extensive experiments have been conducted on MOT17 and MOT20 datasets. The results show that our proposed method outperforms the compared state-of-art tracking methods. For the benefit of the community, our code has been made publicly at https://github.com/mengting2023/LG-Track.*

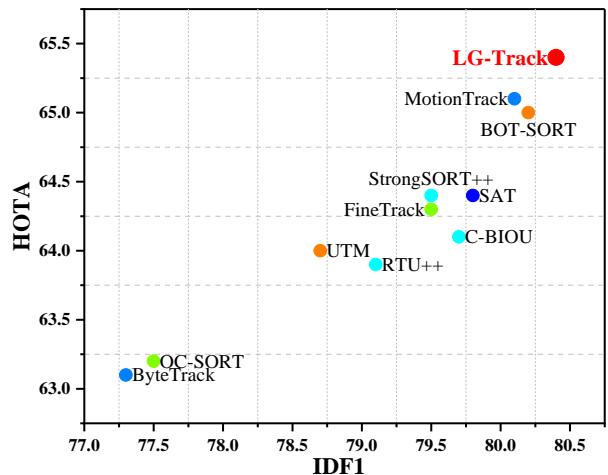

**Fig. 1** HOTA-IDF1 comparisons of state-of-art trackers with our proposed method (LG-Track) based on MOT17 test set.

## 1. Introduction

Multi-object tracking (MOT) is a key technology in computer vision; it is widely applied in autonomous driving, safety monitoring and other fields. Currently, tracking-by-detection (TBD) is a mainstream MOT paradigm, which mainly comprises two sub-tasks: object detection and data association. In recent years, some impressive detectors have been introduced [1]–[9] to classify and localize objects in images, so as to finally derive bounding box coordinates and detection confidence of the objects. Furthermore, Jiang et al. [10] proposed IoU-Net, which predicts the IoU between bounding box and ground-truth as the localization confidence to improve the localization accuracy of object detection. Constant improvement in detection accuracy has contributed greatly to development of the MOT technology for TBD paradigm.

Most existing TBD-based MOT methods [13, 14, 15, 19, 20, 21, 22] are innovatively developed based on data association, with focus on design of a better cost matrix or matching strategy. Yet TBD-based MOT is an integrated task of object detection and data association, the detection performance would directly affect tracking. Object detection, as a step preceding data association, is decisive to the upper bound of data association. Detection may provide more useful information to guide design of association methods and to achieve more reliable tracking.

In light of the foregoing, Zhang et al. proposed ByteTrack [28], which, based on innovation in the junction of detection and association, used detection confidence for hierarchical matching. In this method, trajectories are first matched with high-score detection boxes based on appearance similarity or IoU, while low-score detection boxes are matched with the remaining trajectories by IoU, so as to restore objects in low-score detection boxes. However, we believe that detection confidence can provide more specific and useful information, since for most detectors [4]–[9] detection confidence has integrated classification confidence and localization confidence. Yet in existing TBD-based tracking methods, no consideration is given to localization confidence. Most TBD-based methods assume that objects of low detection confidence are highly occluded, and it is a common practice to disregard such objects or reduce their matching priority. Besides, appearance similarity is not a factor to consider for matching these objects.

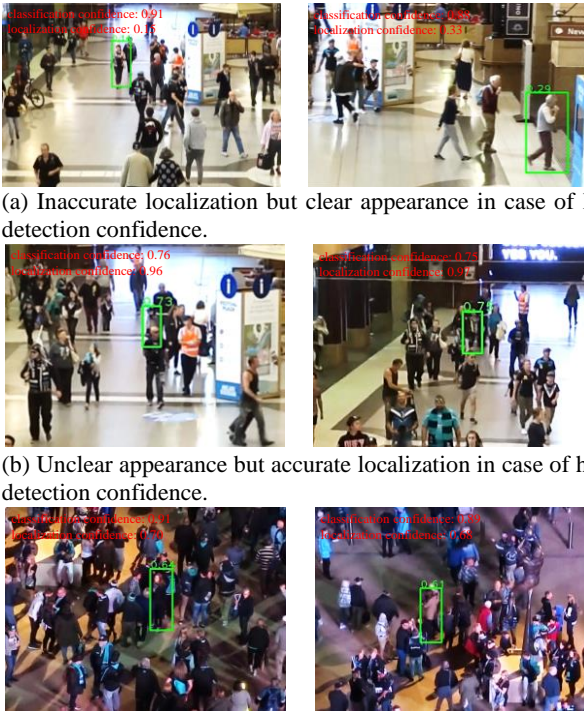

(a) Inaccurate localization but clear appearance in case of low detection confidence.

(b) Unclear appearance but accurate localization in case of high detection confidence.

(c) Inaccurate localization but clear appearance in case of high detection confidence.

**Fig. 2** Different conditions of detection confidence.

It shall be noted that, for the detection confidence integrates classification and localization [4]–[9], objects of low detection confidence may have inaccurate localization but clear appearance, as shown in Fig. 2(a), in which case it is inaccurate localization that leads to the low detection confidence. These objects can be extracted clear appearance features, the appearance similarity should be preferably used for matching and the objects can be better matched than those low-score detection boxes with blurry appearance and inaccurate localization. For high-score detection boxes, no detailed hierarchical matching is designed in existing methods. As a matter of fact, there are high-score objects with inaccurate localization or unclear appearance, which have no advantage in peer competition with those objects with accurate localization and clear appearance. Fig. 2(b) shows high-score detection boxes (with the detection confidence 0.75 and 0.73), in which objects are highly occluded, and thus matching by appearance similarity is inapplicable. Fig. 2(c) shows the objects have high classification confidence (0.91 and 0.89) but low localization confidence (0.7 and 0.68), and thus matching by using a cost matrix of IoU and other motion information may be inaccurate.

To solve the problems above, we propose Localization-Guided Track (LG-Track), which applies localization confidence to MOT tasks for the first time and uses localization confidence and classification confidence to guide selection of a cost matrix, having achieved robust and reliable deep association. Fig. 1 shows comparison of LG-Track and other competitive methods in terms of two key evaluation metrics on MOT17 test set (HOTA and IDF1).

Main contributions of the present study:

● A robust and reliable multi-object tracking method is presented, which makes full use of the detection confidence with consideration given to appearance clarity and localization accuracy, and has a four-level deep association designed to achieve robust data association.

● Localization confidence is applied to MOT tasks for the first time, providing a novel thinking for solving challenges in data association; use of localization confidence and classification confidence makes it possible to select a more appropriate cost matrix.

● Extensive experiments have been conducted on MOT17 and MOT20 datasets. The results show that our proposed method outperforms the compared state-of-art tracking methods.

## 2. Related Works

Generally, the 2D tracking method based on TBD would first proceed with object detection by each frame of video sequences, followed by data association to achieve optimal matching between trajectories and the latest frame detection boxes.

### A. Object Detection

Object detection, as the foundation for realization of multi-object tracking, has developed vigorously in recent years. R-CNN [1]–[3] algorithms are classic two-stage object detection algorithms that divide detection into two stages, i.e. candidates selection; object classification and localization. These algorithms first extract candidate regions of objects through the Region Proposal Network, and then classify and locate the extracted regions through a regression network. Compared with the two-stage algorithms, the one-stage algorithms [4]–[7] reframe object detection as a single regression problem, and use only one network to extract features of whole images for classification and localization, having achieved a good balance between detection accuracy and speed. Jiang et al. [10], finding that CNN-based detectors only output classification confidence and that due to lack of localization confidence, properly localized bounding boxes get deteriorated during iterative regression or suppressed in Non-Maximum Suppression (NMS), put forward IoU-Net, in which IoU between bounding boxes and ground-truth can be predicted to derive the localization confidence, so that the localization accuracy of detectors can be improved. Tian et al. [9] proposed FCOS, a method that adds a single-layer branch to predict the Center-ness of location. Center-ness describes the normalized distance between the center of object and the matched ground-truth; it can reduce the weight of bounding boxes far away from the object center, and makes it possible for these low-quality bounding boxes to be filtered out during the NMS, having prominently improved the detection performance.

From the above, it can be known that introduction of relevant localization score to guide post-detection processing can prominently improve the detection performance.

**B. Data Association**

Data association is the core task of multi-object tracking, which first calculates the cost matrix between trajectories and detection boxes, and then uses different strategies based on the cost matrix for matching.

**Cost Matrix.** Information on motion, location and appearance provides useful clues for data association. Bewley et al. [11] proposed SORT, a method that effectively combines information on motion and location. According to the method, the Kalman filter [12] is first used to predict the location of a trajectory in the current frame, and then the IoU between the predicted box and the detection box is calculated to serve as the cost matrix. Yet IoU simply describes overlapped information between the two boxes, and cannot accurately characterize the center point and the shape, among others. To solve this problem, some methods [13]–[15] are proposed to use GIoU, DIoU and SDIoU as cost matrices for more accurate data association. Wu et al. put forward TraDes [17], which designs a neural network to learn spatio-temporal displacement of an object center in adjacent frames.

When the object is far away or is occluded for some time, its appearance information may play a significant role. Wojke et al. [19] presented DeepSORT, which can extract appearance features from the detection box in images through an independent Re-ID network, and calculate the cosine distance of the feature vectors between trajectories and detection boxes to represent the appearance similarity. Aharon et al. [21] proposed BOT-SORT, which, by fusing motion and appearance information to calculate cost matrices, can achieve better association. Cao et al. [22] put forward OC-SORT, which first seeks to obtain the connecting direction of two observations on trajectories and that of a new observation and the observation corresponding to a trajectory, then calculates to the consistency between the two directions, and weights it with IoU to derive the cost matrix.

**Matching Strategy.** In TBD-based methods, following calculation of the cost matrix, a matching strategy will be used for association between detection boxes and trajectories. This is commonly achieved by Hungary algorithm [23] or greedy algorithm [25]. SORT [11] associates detection boxes and trajectories by once matching. DeepSORT [19] introduced a matching cascade that gives priority to more frequently seen objects to match. Chen et al. [24] proposed MOTDT, which uses the Hierarchical matching; according to this method, appearance similarity is first used for association between all detection boxes and trajectories, and IoU is then used to match the unmatched detection boxes and the unmatched trajectories. Bergmann et al. [25] put forward Tracktor, a method integrating detection and tracking in one module. In this method, a two-stage detector is used; and historical trajectories are used as proposals for regression; finally, NMS is used to associate trajectories and detection boxes. Among the most recent transformer tracking methods, the TrackFormer [27] proposed by Meinhardt et al. realizes data association through track queries, and achieves temporal and spatial tracking of objects in video sequences by an autoregressive fashion. ByteTrack [28] proposed to associate every detection box, and to divide the detection boxes into high-score detection boxes and low-score detection boxes through detection confidence. According to this method, appearance similarity or IoU is first used for matching between high-score detection boxes and trajectories; then IoU is used for matching low-score detection boxes and the unmatched trajectories, to recover objects in low-score detection boxes and filter out the background.

All the methods above focus on how to design better cost matrices or matching strategies. However, the present study concentrates on actual conditions of detection boxes, with consideration given to their localization accuracy and appearance clarity, so as to select more accurate cost matrices and design more reliable matching strategies.

A robust and reliable multi-object tracking method is put forward in this paper. In this method, the localization confidence is applied to MOT for the first time, with consideration given to appearance clarity and localization accuracy of detection boxes, and a four-level deep association mechanism is designed; based on classification confidence and localization confidence, the method allows selection of a more accurate cost matrix. Results of experiments show that the method proposed outperforms SOTA tracking methods on classic datasets (MOT17 and MOT20).

## 3. Proposed Method

The robust and reliable multi-object tracking method proposed herein, as shown in Fig. 3, comprises three parts: input, deep association and output.

Input refers to input of localization confidence, classification confidence and position coordinates from detectors to the method. Deep association has a track management module, in which there are four statuses of trajectories, i.e. New, Tracked, Lost and Removed. The core of deep association is a four-level matching mechanism: (1) detection boxes with high localization confidence and high classification confidence are assigned the highest priority for association with all existing trajectories; (2) after the 1st level of association is done, detection boxes with high localization confidence but low classification confidence are matched with the unmatched trajectories of Tracked and Lost; (3) in the 3rd level of association, detection boxes with low localization confidence and high classification confidence are matched with the remaining trajectories of Tracked and Lost; (4) finally, the remaining unmatched Tracked trajectories are matched with detection boxes with low localization confidence and low classification confidence.

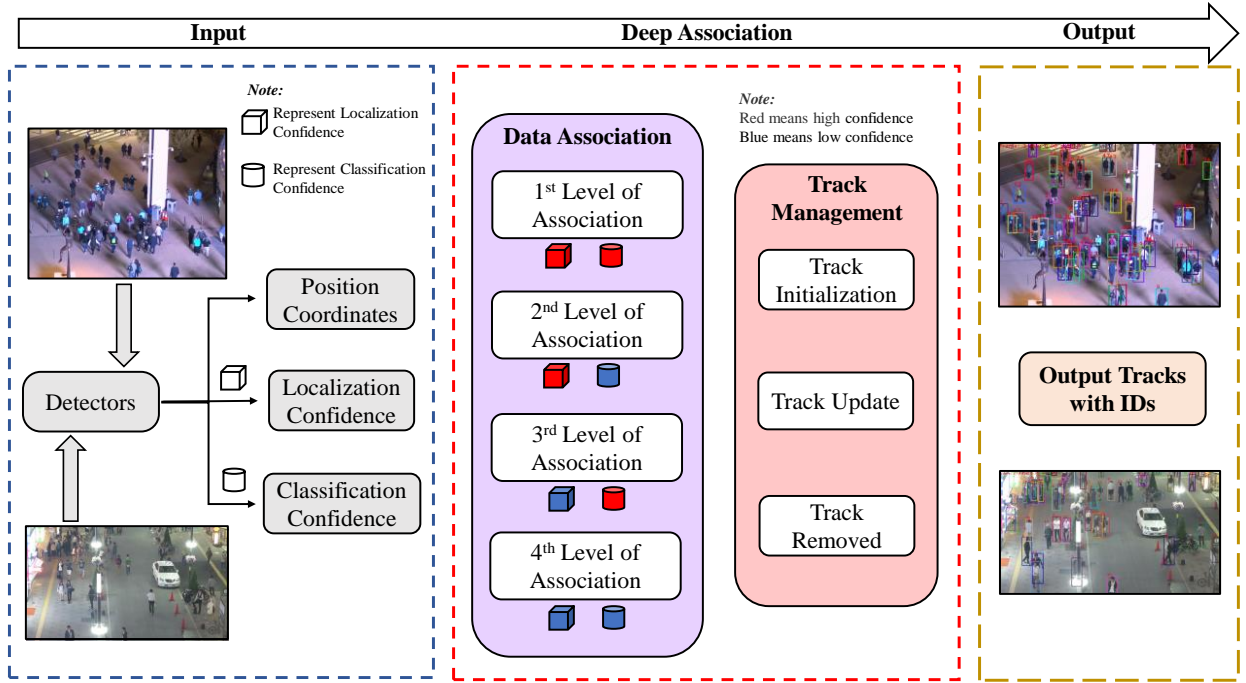

**Fig. 3** Overall structure of the proposed method.

### A. 2D Detector

The method proposed herein requires output of localization confidence and classification confidence from detectors. For CNN-based detectors [1]–[3] that simply output classification confidence, IoU-Net [10] may be used to derive the localization confidence. For YOLO series detectors [4]–[8] that output detection confidence integrating localization and classification, the localization confidence and the classification confidence can be obtained simply by separating the detection confidence. In FCOS [9], in which Centerness plays a similar role with the localization confidence.

### B. Deep Association

**Data Association.** A multi-level deep association mechanism is proposed in this paper, and the mechanism mainly has the following contributions: (1) full consideration is given to the localization accuracy and appearance clarity of detection boxes, and a proper cost matrix is selected; (2) objects with inaccurate localization or ambiguous appearance in high-score detection boxes and objects with accurate localization or clear appearance in low-score detection boxes are separated and assigned with more reasonable priority in matching. The deep association mechanism in this paper includes four-level data association, which is described in detail as follows.

1st Level of Association: In the 1st level of association, all existing trajectories are matched with detection boxes of high localization confidence and high classification confidence, and the following cost matrix $C_1$ that based on motion information is used for matching:

$$C_1 = C_{\text{iou}} \cdot d_l \quad (1)$$

$$C_{\text{iou}}(i,j) = \frac{d_i \cap t_j}{d_i \cup t_j} \quad (2)$$

where, $d_i$ refers to the $i^{\text{th}}$ detection box; $t_j$ refers to the $j^{th}$ trajectory; $C_{iou}(i,j)$ refers to the IoU between the $i^{th}$ detection box and the $j^{th}$ trajectory; $d_l$ refers to the localization confidence of detection boxes.

In scenes congested with pedestrians, some detection boxes, which may be occluded by each other, may have approximate IoU with the same trajectory, which may cause mismatch of the right detection boxes. To solve this problem, we multiply the localization confidence of corresponding detection boxes by the IoU in Formula (1) to reduce the cost matrices corresponding to detection boxes of low localization confidence, so as to make the cost matrix more accurate and reasonable. Upon completion of the 1st level of association, three possible association results may be obtained: matched trajectories $t_{m1}$, unmatched trajectories $t_{u1}$ and unmatched detection boxes $d_{u1}$. New trajectories in $t_{u1}$ will turn into removed trajectories. The remainder of $t_{u1}$ and $d_{u1}$ will move to the 2nd level of association.

2nd Level of Association: In the 2nd level of association, detection boxes with high localization confidence but low classification confidence and $d_{u1}$ are matched with $t_{u1}$. Since detection boxes to be matched at this level are accurately located with unclear appearance, the cost matrix used in association at this level is the same with that used in the 1st level of association. Similarly, matching at this level will also produce three possible association results: matched trajectories $t_{m2}$, unmatched trajectories $t_{u2}$ and unmatched detection boxes $d_{u2}$. $t_{u2}$ will move to the 3rd level of association, while $t_{m2}$ will move to the final level of association.

3rd Level of Association: In the 3rd level of association, detection boxes of low localization confidence but high classification confidence are matched with $t_{u2}$. Since the detection boxes to be matched at this level have clear appearance but inaccurate localization, appearance information is used in calculation of the cost matrix $C_3$:

$$C_3 = C_{cos} \cdot d_s \quad (3)$$

where, $C_{cos}$ refers to the similarity between the detection box and the trajectory in terms of appearance, and is expressed by using the cosine distance; $d_s$ is the detection confidence of detection boxes, which has integrated the localization confidence and the classification confidence.

In this work, we use the detection confidence multiplied by $C_{cos}$ to derive the cost matrix $C_3$, because appearance features are extracted from corresponding detection box in the image, while the localization accuracy of the detection box may affect accuracy of the appearance features. Matching at this level also produces three possible association results: matched trajectories $t_{m3}$, unmatched trajectories $t_{u3}$ and unmatched detection boxes $d_{u3}$. Both $t_{u3}$ and $d_{u3}$ will move to the final level of association.

4th Level of Association: In the 4th level of association, detection boxes of low localization confidence and low classification confidence and all unmatched detection boxes above are matched with Tracked trajectories in $t_{u3}$. Matching at this level is performed by using the following cost matrix $C_4$:

$$C_4 = \alpha C_1 + (1 - \alpha)C_3 \quad (4)$$

where, $C_1$ and $C_3$ are cost matrices respectively used in the 1st level of association and the 3rd level of association. Since matching at this level mainly involves objects with inaccurate localization and ambiguous appearance, we simultaneously consider motion and appearance information of the objects by weighting motion and appearance similarity to derive $C_4$.

Matching at this level will finally produce three possible association results: matched trajectories $t_{m4}$, unmatched trajectories $t_{u4}$ and unmatched detection boxes $d_{u4}$. In $d_{u4}$, objects with the detection confidence of higher than $S_{low}$ are initialized to New trajectories, and low-score detections are deleted. $d_{u4}$ will update the trajectory status following the track management module.

**Track Management.** It is known that good track management can avoid false negative and false positive. In the method proposed herein, the track management strategy in [21] is adopted to handle trajectories in entry and exit scenes. For trajectories, there are four statuses, i.e. New, Tracked, Lost and Removed. The status, New, is assigned to initialized trajectories of objects that have the detection confidence of higher than $S_{low}$. After successful matching by $T_{max}$ frames, the trajectories will have their status change to Tracked. When trajectories in the Tracked status fail to be matched for consecutively $L_{max}$ frames, the status will change to Removed. In addition, the New trajectories that have one frame fails to be matched will also have the status change to Removed.

The results below prove that the deep association mechanism above can achieve superior tracking performance, and outperforms the compared SOTA tracking methods.

## 4. Experiments

In this section, comparative experimental results are demonstrated to illustrate the effectiveness of the method proposed herein.

### A. Experimental Settings

**Datasets.** We compare LG-Track with state-of-the-art tracking methods on the test set of MOT17 [29] and MOT20 [30] under the "private detector" protocol. For fair comparison with these methods, we used the publicly available YOLOX [8] trained by ByteTrack [28] as the detector. To extract appearance information of objects, we used the FastReID [31] SBS-50 model trained by BOT-SORT [21].

**Metrics.** The method proposed in this paper is evaluated by the two commonly used metrics, i.e. CLEAR [32] and HOTA [33]. CLEAR involves several sub-metrics, mainly including Multi-Object Tracking Accuracy (MOTA), Multi-Object Tracking Precision (MOTP), ID Switch (IDs) and IDF1. HOTA takes into account both detection and association performance, and it also has several sub-metrics, such as detection accuracy (DetA) and association accuracy (AssA).

**Implementation Details.** For LG-Track, the localization confidence threshold is 0.55 and the classification confidence threshold is 0.75, unless otherwise specified. In the data association module, the cost matrix thresholds for the four levels of matching are 0.65, 0.65, 0.5 and 0.55, respectively. For trajectories identified as Lost, we keep them for 30 frames to avoid assignment of new IDs to them upon their reappearance.

Table 1. Comparison of the state-of-the-art methods under the "private detector" protocol on MOT17 test set. In methods with blue shading, YOLOX [8] is used. Data marked in red and bold are the best of corresponding metrics; data marked in blue and bold are the second of corresponding metrics.

| Method | Published Year | HOTA (↑) | MOTA (↑) | IDF1 (↑) | AssA (↑) | DetA (↑) | IDSW (↑) |
|---|---|---|---|---|---|---|---|
| MOTR [26] | ECCV (2022) | 57.8 | 73.4 | 68.6 | 55.7 | 60.3 | 2439 |
| TrackFormer [27] | CVPR (2022) | 57.3 | 74.1 | 68.0 | 54.1 | 60.9 | 2829 |
| RelationTrack [34] | TMM (2022) | 61.0 | 73.8 | 74.7 | 61.5 | 60.6 | 1374 |
| MAATrack [35] | WACVw (2022) | 62.0 | 79.4 | 75.9 | 60.2 | 64.2 | 1452 |
| RTU++ [37] | TIP (2022) | 63.9 | 79.5 | 79.1 | 63.7 | 64.5 | 1302 |
| SGT [38] | WACV (2023) | ---- | 76.4 | 72.8 | ---- | ---- | 4101 |
| UTM [40] | CVPR (2023) | 64.0 | **81.8** | 78.7 | 62.5 | **65.9** | 1431 |
| BOT-SORT [21] | ArXiv (2022) | 65.0 | 80.5 | **80.2** | **65.5** | 64.9 | 1212 |
| ByteTrack [28] | ECCV (2022) | 63.1 | 80.3 | 77.3 | 62.0 | 64.5 | 2196 |
| SAT [36] | MM (2022) | 64.4 | 80.0 | 79.8 | 64.4 | 64.8 | 1356 |
| StrongSORT++ [20] | TMM (2023) | 64.4 | 79.6 | 79.5 | 64.4 | 64.6 | 1194 |
| OC-SORT [22] | CVPR (2023) | 63.2 | 78.0 | 77.5 | 63.2 | ---- | 1950 |
| C-BIOU [39] | WACV (2023) | 64.1 | 81.1 | 79.7 | 63.7 | 64.8 | 1455 |
| FineTrack [41] | CVPR (2023) | 64.3 | 80.0 | 79.5 | 64.5 | 64.5 | 1272 |
| MotionTrack [42] | CVPR (2023) | **65.1** | 81.1 | 80.1 | 65.1 | 65.4 | **1140** |
| **LG-Track** | ---- | **65.4** | 81.4 | **80.4** | 65.4 | 65.6 | **1125** |

Table 2. Comparison of the state-of-the-art methods under the "private detector" protocol on MOT20 test set. In methods with blue shading, YOLOX [8] is used. Data marked in red and bold are the best of corresponding metrics; data marked in blue and bold are the second of corresponding metrics.

| Method | Published Year | HOTA (↑) | MOTA (↑) | IDF1 (↑) | AssA (↑) | DetA (↑) | IDSW (↑) |
|---|---|---|---|---|---|---|---|
| TrackFormer [27] | CVPR (2022) | 54.7 | 68.6 | 65.7 | 53.0 | 56.7 | 1532 |
| Relation Track [34] | TMM (2022) | 56.5 | 67.2 | 70.5 | 56.4 | 56.8 | 4243 |
| MAATrack [35] | WACVw (2022) | 57.3 | 73.9 | 71.2 | 55.1 | 59.7 | 1331 |
| RTU++ [37] | TIP (2022) | 62.8 | 76.5 | 76.8 | 62.6 | 63.1 | 971 |
| SGT [38] | WACV (2023) | ---- | 72.8 | 70.6 | ---- | ---- | 2474 |
| UTM [40] | CVPR (2023) | 62.5 | **78.2** | 76.9 | 61.4 | 63.7 | 1228 |
| BOT-SORT [21] | ArXiv (2022) | **63.3** | 77.8 | **77.5** | 62.9 | **64.0** | 1313 |
| ByteTrack [28] | ECCV (2022) | 61.3 | 77.8 | 75.2 | 59.6 | 63.4 | 1223 |
| SAT [36] | MM (2022) | 62.6 | 75.0 | 76.6 | **63.2** | 62.1 | 816 |
| OC-SORT [22] | CVPR (2023) | 62.1 | 75.5 | 75.9 | 62.0 | ---- | **913** |
| StrongSORT++ [20] | TMM (2023) | 62.6 | 73.8 | 77.0 | **64.0** | 61.3 | **770** |
| MotionTrack [42] | CVPR (2023) | 62.8 | **78.0** | 76.5 | 61.8 | **64.0** | 1165 |
| **LG-Track** | ---- | **63.4** | 77.8 | **77.4** | 62.9 | **64.1** | 1161 |

## B. Comparison with the State-of-the-Art Methods

**Quantitative Evaluation.** Based on MOT17 [29] and MOT20 [30] test sets, we compared the LG-Track proposed in this paper with other SOTA tracking methods in terms of major tracking metrics. The results are shown in Table 1 and Table 2. It should be noted that in Table 1 and Table 2, the methods with blue shading use YOLOX [8] as the detector. Data marked in red and bold are the best of corresponding metrics, while data marked in blue and bold are the second of corresponding metrics.

Table 1 gives the results of comparison between LG-Track and 15 SOTA tracking methods based on the MOT17 test set. In terms of the key evaluation metrics, HOTA, LG-Track outperformed the 15 competitive methods by achieving the best results (65.4%); in addition, it also achieved the highest IDF1 (80.4%) and the lowest IDSW (1125). Results based on the MOT20 test set are shown in Table 2. Compared with 12 SOTA tracking methods, LG-Track achieved the highest HOTA (63.4%) and DetA (64.1%). It shall be pointed out that though LG-Track did not achieved the highest MOTA among the competitive methods, a good balance between HOTA and MOTA was achieved with LG-Track (MOT17: HOTA 65.4%, MOTA 81.4%; MOT20: HOTA 63.4%, MOTA 77.8%), relative to UTM (MOT17: HOTA 64%, MOTA 81.8%; MOT20: HOTA 62.5%, MOTA 78.2%) that achieved the highest MOTA. This suggests that the LG-Track proposed herein provides the best overall tracking performance among the competitive methods.

**Qualitative Evaluation.** Fig. 4 shows the results of visualized comparison between LG-Track proposed in this paper and ByteTrack [28], OC-SORT [22] and BOT-SORT [21] in MOT17-02 and MOT17-10 scenes. It can be seen that in the two scenes above, the three trackers in comparison had ID switch to different extents, while LG-Track realized robust and reliable tracking without ID switch. In frames 97 and 98 of MOT17-10, it is found that the pedestrian with ID 4 has a very close IoU with the pedestrian just entering the image, and appearance features of the two objects extracted from the detection boxes are highly similar due to high occlusion. The two objects were taken wrongly as the same object in tracking by the three comparative methods; yet they were reliably tracked by LG-Track as a new ID was assigned to the pedestrian just entering the image.

Table 3. Comparison of results by using different cost matrices based on MOT17 training set. Data marked in red and bold are the best of corresponding metrics.

| $2^{nd}$ | $3^{rd}$ | HOTA(↑) | AssA(↑) | MOTA(↑) | IDF1(↑) |
|---|---|---|---|---|---|
| $C_1$ | $C_1$ | 77.21 | 75.32 | **90.74** | 85.79 |
| $C_3$ | $C_3$ | 77.61 | 76.16 | 90.45 | 86.49 |
| $C_1$ | $C_3$ | **77.72** | **76.35** | 90.68 | **86.54** |

Table 4. Comparison of results when IoU is penalized by different confidence, based on MOT17 training set. Data marked in red and bold are the best of corresponding metrics. $IoU \cdot d_s$ means IoU is fused with the detection confidence. $IoU \cdot d_c$ means IoU is fused with the classification confidence. $IoU \cdot d_l$ means IoU is fused with the localization confidence.

| | HOTA(↑) | AssA(↑) | MOTA(↑) | IDF1(↑) |
|---|---|---|---|---|
| $IoU \cdot d_s$ | 77.31 | 75.62 | 90.44 | 86.15 |
| $IoU \cdot d_c$ | 77.24 | 75.23 | 90.59 | 85.75 |
| $IoU \cdot d_l$ | **77.72** | **76.35** | **90.68** | **86.54** |

Table 5. Effect of multi-level deep association mechanism, based on MOT17 training set. Data marked in red and bold are the best of corresponding metrics.

| $1^{st}$ | $2^{nd}$ | $3^{rd}$ | $4^{th}$ | HOTA(↑) | AssA(↑) | MOTA(↑) | IDF1(↑) |
|---|---|---|---|---|---|---|---|
| √ | × | × | × | 77.00 | 75.48 | 89.74 | 86.16 |
| √ | × | √ | √ | 77.49 | **76.41** | 89.76 | 86.50 |
| √ | √ | × | √ | 77.43 | 75.83 | 90.67 | 86.46 |
| √ | √ | √ | × | 77.38 | 75.75 | 90.65 | 86.39 |
| √ | √ | √ | √ | **77.72** | 76.35 | **90.68** | **86.54** |

## C. Ablation Study

In this section, we firstly investigate how localization confidence and classification confidence can be used to guide the selection of cost matrix. Then, we look into how different confidences (i.e. localization confidence, classification confidence, and detection confidence) should be fused with IoU to achieve superior tracking performance. Lastly, the effects of different association levels on the tracking performance are investigated. The MOT17 training set was used for our ablation study.

**Selection of cost matrix.** In the $2^{nd}$ level of association (detection boxes with high localization confidence but low classification confidence) of the multi-level deep association mechanism proposed in this paper, the cost matrix $C_1$ based on motion information was used; in the $3^{rd}$ level of association (detection boxes of low localization confidence but high classification confidence), the cost matrix $C_3$ based on appearance information was selected. This ablation study adopted the cost matrix $C_1$ or the cost matrix $C_3$ at both the $2^{nd}$ level of association and the $3^{rd}$ level of association. The results are shown in Table 3, which indicate that use of localization confidence and classification confidence to guide selection of cost matrixes can achieve better tracking performance (higher HOTA, AssA and IDF1).

**Different confidence fusing with IoU.** The localization confidence multiplied with IoU is used to derive the motion information-based cost matrix $C_1$ in the proposed method. In this ablation study, the localization confidence, the classification confidence and the combination of both (i.e. the detection confidence) were respectively multiplied with IoU to derive the cost matrix $C_1$. The results on MOT17 training set are shown in Table 4, which indicates that use of localization confidence fusing with IoU can achieve better tracking performance.

**Effects of multi-level deep association mechanism.** In this ablation study, we look into the effects of different association levels on tracking performance. We conducted individual tests by removing the $2^{nd}$, the $3^{rd}$, the $4^{th}$ level of association separately, and removing these three levels altogether. The results on MOT17 training set are shown in Table 5. We see that the proposed multi-level deep association mechanism shows the best overall tracking performance, in comparison with other association strategies with fewer association levels.

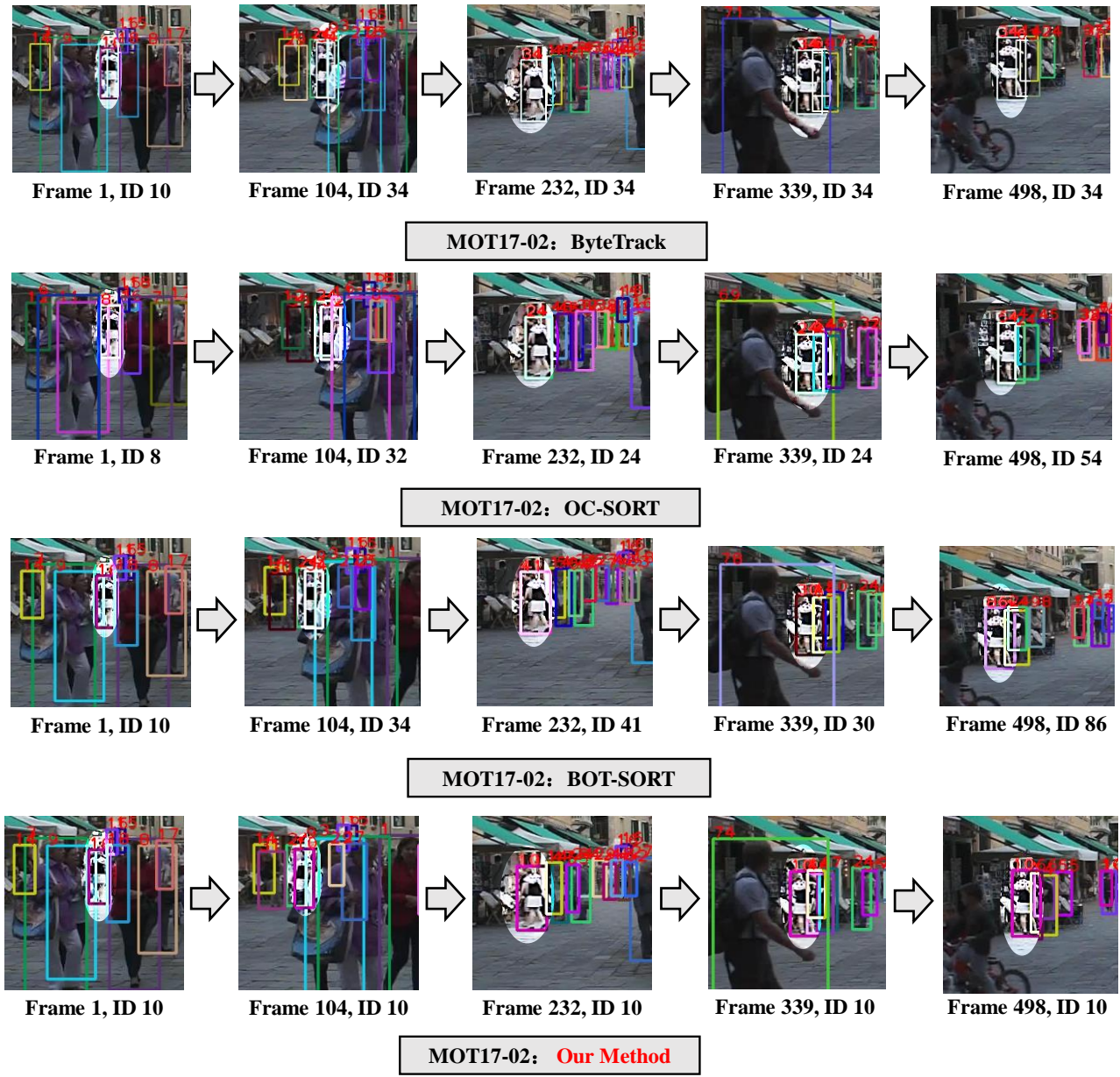

(a) Results of comparison with four methods based on MOT17-02.

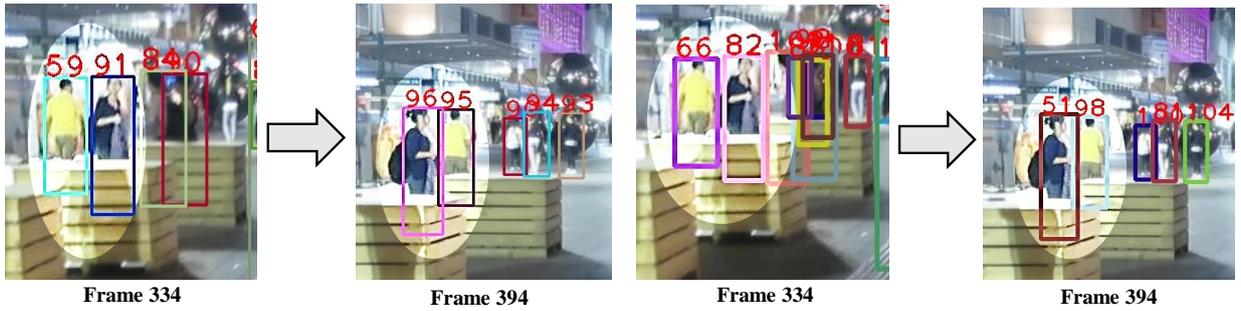
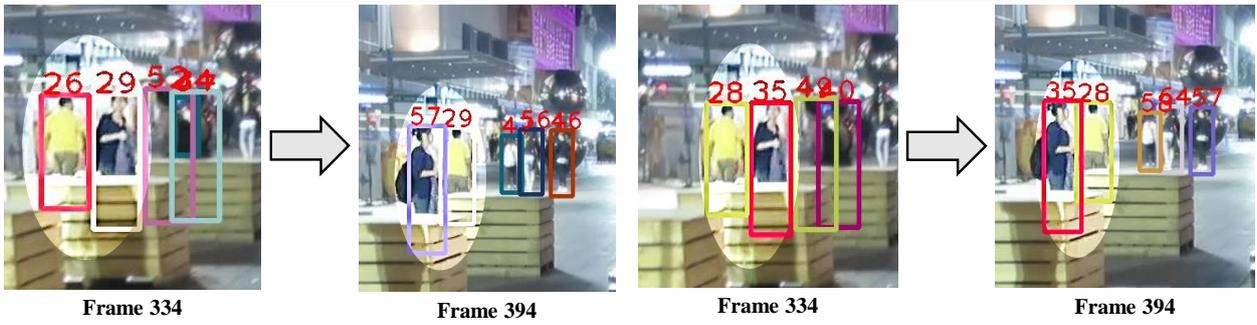
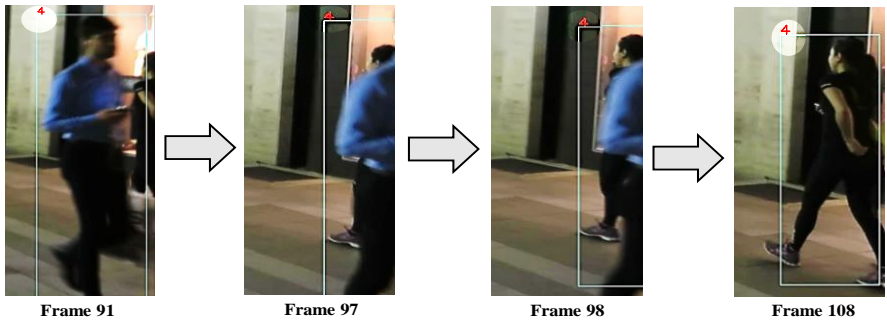
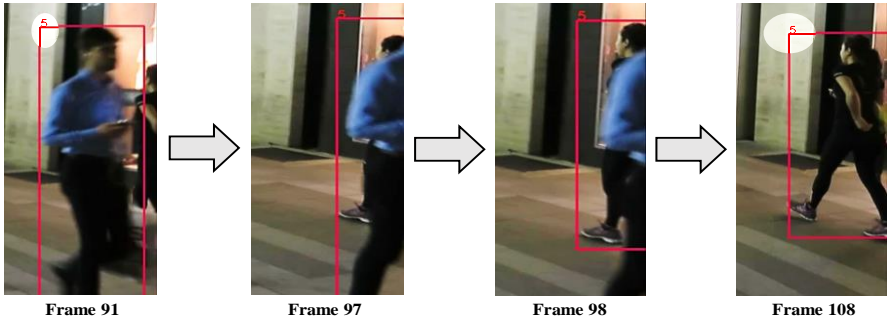
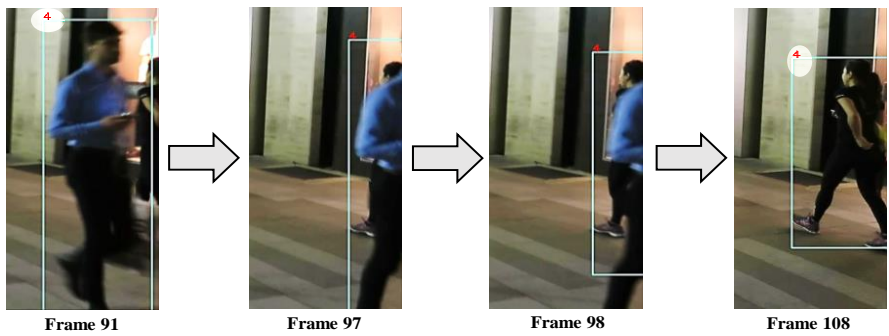

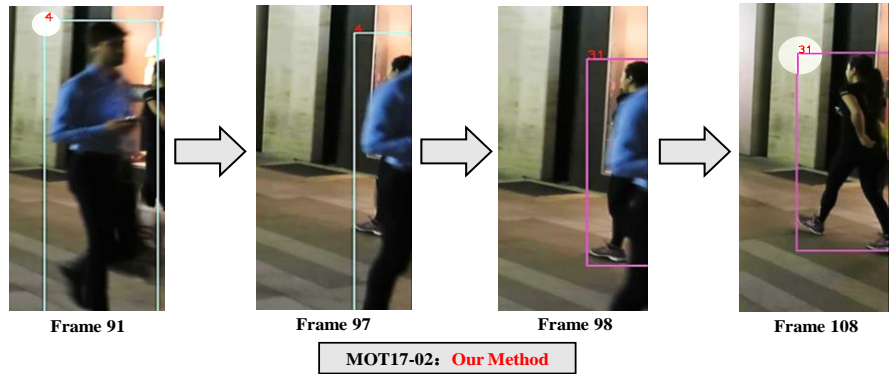

Frame 91　Frame 97　Frame 98　Frame 108

MOT17-02： **Our Method**

(b) Comparison with four methods based on MOT17-10.

**Fig. 4** Comparison of the proposed method with ByteTrack, OC-SORT and BOT-SORT in terms of visualized results based on MOT17-02 and MOT17-10 scenes.

# 5. Conclusion

We propose a strong tracker, which makes full use of detection confidence, with consideration given to appearance clarity and localization accuracy, and has a four-level deep association mechanism; in this tracker, localization confidence is applied to MOT tasks for the first time, and a more appropriate cost matrix is selected based on localization confidence and classification confidence. Compared with other SOTA trackers, the tracker proposed herein achieves the best overall tracking performance on MOT17 and MOT20 datasets.

## Acknowledgments

This work was supported by the Chongqing Technology Innovation and Application Development Project under Grant CSTB2022TIAD-DEX0013.